\pdfoutput=1

\documentclass[11pt]{article}

\usepackage[]{emnlp2021}

\usepackage{times}
\usepackage{latexsym}

\usepackage[T1]{fontenc}

\usepackage[utf8]{inputenc}

\usepackage{microtype}

\usepackage{newtxmath}
\usepackage{amsmath}
\usepackage{bm}
\usepackage{makecell}
\usepackage{caption}
\usepackage{array}
\usepackage{graphicx, subfig}
\usepackage{multirow}
\usepackage{multicol}
\usepackage{booktabs}
\usepackage{float}
\usepackage{tabularx}
\usepackage{braket}

\usepackage{paralist}
\usepackage{paralist}

\usepackage{xcolor}

\newenvironment{shrinkeq}[1]
{ \bgroup
  \addtolength\abovedisplayshortskip{#1}
  \addtolength\abovedisplayskip{#1}
  \addtolength\belowdisplayshortskip{#1}
  \addtolength\belowdisplayskip{#1}}
{\egroup\ignorespacesafterend}

\title{TimeTraveler: Reinforcement Learning for Temporal Knowledge \\ Graph Forecasting}

\author{
Haohai Sun\textsuperscript{1}\footnotemark[1] \quad 
Jialun Zhong\textsuperscript{1}\footnotemark[1] \quad 
Yunpu Ma\textsuperscript{2}\footnotemark[1] \quad 
Zhen Han\textsuperscript{2,3}\footnotemark[1] \quad
Kun He\textsuperscript{1}\footnotemark[2] \\
\textsuperscript{\rm 1} School of Computer Science and Technology, \\ Huazhong University of Science and Technology \\
\textsuperscript{\rm 2} Institute of Informatics, LMU Munich 
\textsuperscript{\rm 3} Corporate Technology, Siemens AG \\
\texttt{\{haohais, zhongjl\}@hust.edu.cn} \\
\texttt{cognitive.yunpu@gmail.com, zhen.han@campus.lmu.de} \\
\texttt{brooklet60@hust.edu.cn} \\
}

\begin{document}
\maketitle{

\renewcommand{\thefootnote}{\fnsymbol{footnote}}
\setcounter{footnote}{0}

\footnotetext[1]{Equal Contribution.}
\footnotetext[2]{Corresponding author.}
}

\begin{abstract}
Temporal knowledge graph (TKG) reasoning is a crucial task that has gained increasing research interest in recent years.
Most existing methods focus on reasoning at past timestamps to complete the missing facts,
and there are only a few works of reasoning on known TKGs to forecast future facts. 
Compared with the completion task, the forecasting task is more difficult and faces two main challenges: 
(1) how to effectively model the time information to handle future timestamps? 
(2) how to make inductive inference to handle previously unseen entities that emerge over time?
To address these challenges, we propose the first reinforcement learning method for forecasting. 
Specifically, the agent travels on historical knowledge graph snapshots to search for the answer.
Our method defines a relative time encoding function to capture the timespan information, and we design a novel time-shaped reward based on Dirichlet distribution to guide the model learning. 
Furthermore, we propose a novel representation method for unseen entities to improve the inductive inference ability of the model.
We evaluate our method for this link prediction task at future timestamps. 
Extensive experiments on four benchmark datasets demonstrate substantial performance improvement meanwhile with higher explainability, less calculation, and fewer parameters when compared with existing state-of-the-art methods. 

\end{abstract}

\renewcommand{\thefootnote}{\arabic{footnote}}
\setcounter{footnote}{0}

\section{Introduction}
\label{sec:01intro}
Storing a wealth of human knowledge and facts, Knowledge Graphs (KGs) are widely used for many downstream Artificial Intelligence (AI) applications, such as recommendation systems~\cite{guo2020survey}, dialogue generation~\cite{moon2019opendialkg}, and question answering~\cite{zhang2018variational}. 
KGs store facts in the form of triples $(e_s, r, e_o)$, i.e. (subject entity, predicate/relation, object entity), such as \textit{(LeBron\_James, plays\_for, Cleveland\_Cavaliers)}. 
Each triple corresponds to a labeled edge of a multi-relational directed graph. 
However, facts constantly change over time. To reflect the timeliness of facts, Temporal Knowledge Graphs (TKGs) additionally associate each triple with a timestamp $(e_s, r, e_o, t)$, e.g., \textit{(LeBron\_James, plays\_for, Cleveland\_Cavaliers, 2014-2018)}. Usually, we represent a TKG as a sequence of static KG snapshots. 

TKG reasoning is a process of inferring new facts from known facts, which can be divided into two types, interpolation and extrapolation. Most existing methods~\cite{DBLP:conf/coling/JiangLGSCLS16, dasgupta2018hyte, DBLP:conf/aaai/GoelKBP20, DBLP:conf/emnlp/WuCCH20} focus on interpolated TKG reasoning to complete the missing facts at past timestamps. In contrast, extrapolated TKG reasoning focuses on forecasting future facts (events). In this work, we focus on extrapolated TKG reasoning by designing a model for the link prediction at future timestamps. 
E.g., ``which team LeBron James will play for in 2022?'' can be seen as a query of link prediction at a future timestamp: \textit{(LeBron\_James, plays\_for, ?, 2022)}. 

Compared with the interpolation task, there are two challenges for extrapolation. (1) Unseen timestamps: the timestamps of facts to be forecast do not exist in the training set. (2) Unseen entities: new entities may emerge over time, and the facts to be predicted may contain previously unseen entities. Hence, the interpolation methods can not treat the extrapolation task.

The recent extrapolation method RE-NET~\cite{jin2020Renet} uses Recurrent Neural Network (RNN) to capture temporally adjacent facts information to predict future facts.  CyGNet~\cite{zhu-etal-2021-cygnet} focuses on the repeated pattern to count the frequency of similar facts in history. However, these methods only use the random vectors to represent the previously unseen entities and view the link prediction task as a multi-class classification task, causing it unable to handle the second challenge. Moreover, they cannot explicitly indicate the impact of historical facts on predicted facts.
%


Inspired by path-based methods ~\cite{DBLP:conf/iclr/DasDZVDKSM18, DBLP:conf/emnlp/LinSX18} for static KGs, we propose a new temporal-path-based reinforcement learning (RL) model for extrapolated TKG reasoning. We call our agent the ``\textbf{TI}me \textbf{T}ravel\textbf{er}'' (\textbf{TITer}), which travels on the historical KG snapshots to find answers for future queries. 
TITer starts from the query subject node, sequentially transfers to a new node based on temporal facts related to the current node, and is expected to stop at the answer node. 
To handle the unseen-timestamp challenge, TITer uses a relative time encoding function to capture the time information when making a decision. We further design a novel time-shaped reward based on Dirichlet distribution to guide the model to capture the time information. 
To tackle the unseen entities, we introduce a temporal-path-based framework and propose a new representation mechanism for unseen entities, termed the Inductive Mean (IM) representation, so as to improve the inductive reasoning ability of the model. 


Our main contributions are as follows:
\begin{itemize}

\item This is the first temporal-path-based reinforcement learning model for extrapolated TKG reasoning, which is explainable and can handle unseen timestamps and unseen entities. 

\item We propose a new method to model the time information. We utilize a relative time encoding function for the agent to capture the time information and use a time-shaped reward to guide the model learning.

\item We propose a novel representation mechanism for unseen entities, which leverages query and trained entity embeddings to represent untrained (unseen) entities. This can stably improve the performance for 
inductive inference without increasing the computational cost. 


\item Extensive experiments indicate that our model substantially outperforms existing methods with less calculation and fewer parameters.
\end{itemize}

\section{Related Work} 
\label{sec:relatedWork}
\subsection{Static Knowledge Graph Reasoning}
Embedding-based methods represent entities and relations as low-dimensional embeddings in different representation spaces, such as Euclidean space~\cite{nickel2011three, bordes2013translating}, complex vector space~\cite{trouillon2016complex, DBLP:conf/iclr/SunDNT19}, and manifold space~\cite{DBLP:conf/acl/ChamiWJSRR20}. These methods predict missing facts by scoring candidate facts based on entity and relation embeddings. 
Other works use deep learning models to encode the embeddings, such as Convolution Neural Network (CNN)~\cite{dettmers2018convolutional, vashishth2020interacte} to obtain deeper semantics, or Graph Neural Network (GNN)~\cite{schlichtkrull2018modeling, nathani-etal-2019-learning, zhang2020relational} to encode multi-hop structural information. 

Besides, path-based methods are also widely used in KG reasoning. \citet{lin-etal-2015-modeling} and \citet{DBLP:conf/icml/GuoSH19} use RNN to compose the implications of paths. Reinforcement learning methods~\cite{DBLP:conf/emnlp/XiongHW17, DBLP:conf/iclr/DasDZVDKSM18, DBLP:conf/emnlp/LinSX18} view the task as a Markov decision process (MDP) to find paths between entity pairs, which are more explanatory 
than embedding-based methods.

\subsection{Temporal Knowledge Graph Reasoning}
A considerable amount of works extend static KG models to the temporal domain. These models redesign embedding modules and score functions related to time~\cite{jiang2016towards, dasgupta2018hyte, DBLP:conf/aaai/GoelKBP20, DBLP:conf/iclr/LacroixOU20, DBLP:conf/emnlp/HanCMT20}. Some works leverage message-passing networks to capture graph snapshot neighborhood information~\cite{DBLP:conf/emnlp/WuCCH20, jung2020t}. These works are designed for interpolation.

For extrapolation, Know-Evolve~\cite{trivedi2017know} and GHNN~\cite{han2020graph} use temporal point process to model facts evolved in the continuous time domain. Additionally, TANGO~\cite{ding2021temporal} explores the neural ordinary differential equation to build a continuous-time model. RE-NET~\cite{jin2020Renet} considers the multi-hop structural information of snapshot graphs and uses RNN to model entity interactions at different times. CyGNet~\cite{zhu-etal-2021-cygnet} finds that many facts often show a repeated pattern and make reference to known facts in history. These approaches lack to explain their predictions and cannot handle the previously unseen entities. Explanatory model xERTE~\cite{han2021explainable} uses a subgraph sampling technique to build an inference graph. Although the representation method that refers to GraphSAGE~\cite{DBLP:conf/nips/HamiltonYL17} makes it possible to deal with unseen nodes, the continuous expansion of inference graphs also severely restricts the inference speed.

\section{Methodology}
\label{sec:method}
\begin{figure*}
    \centering
    \includegraphics[width=\linewidth]{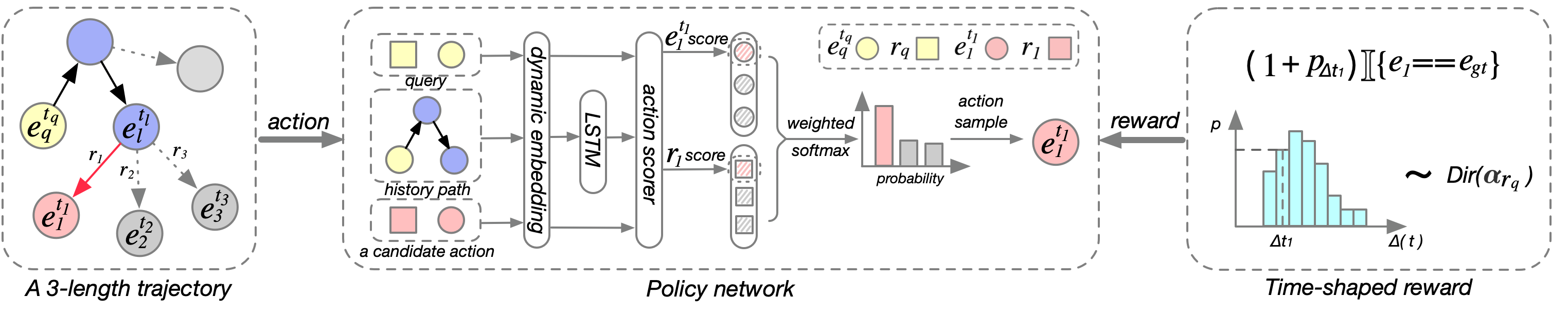}
    \caption{Overview of TITer. Given a query $(e_q, r_q, ?~(e_{gt}), t_q)$, TITer starts from node $e_q^{t_q}$. At each step, TITer samples an outgoing edge and traverses to a new node according to $\pi_{\theta}$ (policy network). We use the last step of the search as an example. $e_l^{t_l}$ is the current node. Illustration of policy network provides the process for scoring one of the candidate actions $(r_1, e_1, t_1)$. TITer samples an action based on the transition probability calculated from all candidate scores.
    When the search is completed, the time-shaped reward function will give the agent a reward based on the estimated Dirichlet distribution $Dir(\bm{\alpha_{r_q}})$. 
    }
    \vspace{-0.5em}
    \label{policy network illustration}
\end{figure*}

Analogizing to the previous work on KGs~\cite{DBLP:conf/iclr/DasDZVDKSM18}, we frame the RL formulation as ``walk-based query-answering'' on a temporal graph: the agent starts from the source node (subject entity of the query) and sequentially selects outgoing edges to traverse to new nodes until reaching a target. In this section, we first define our task, and then describe the reinforcement learning framework and how we incorporate the time information into the on-policy reinforcement learning model. The optimization strategy and the inductive mean representation method for previously unseen entities are provided in the end. Figure \ref{policy network illustration} is the overview of our model.

\subsection{Task Definition} \label{formulation}
Here we formally define the task of extrapolation in a TKG. 
Let $\mathcal{E}$, $\mathcal{R}$, $\mathcal{T}$, and $\mathcal{F}$ denote the sets of entities, relations, timestamps, and facts, respectively. 
A fact in a TKG can be represented in the form of a quadruple $(e_s, r, e_o, t)$, where $r \in \mathcal{R}$ is a directed labeled edge between a subject entity $e_s \in \mathcal{E}$ and an object entity $e_o \in \mathcal{E}$ at time $t \in \mathcal{T}$. We can represent a TKG by the graph snapshots over time. 
A TKG can be described as $\mathcal{G}_{(1, T)} = \{\mathcal{G}_1, \mathcal{G}_2,...,\mathcal{G}_T\}
$, where $\mathcal{G}_t = \{\mathcal{E}_t, \mathcal{R}, \mathcal{F}_t\}$ is a multi-relational directed TKG snapshot, and $\mathcal{E}_t$ and $\mathcal{F}_t$ denote entities and facts that exist at time $t$. In order to distinguish the graph nodes at different times, we let a node be a two-tuple with entity and timestamp: $e^t_i = (e_i, t)
$. Thus, a fact (or event) $(e_s, r, e_o, t)$ can also be seen as an edge from source node $e_s^t$ to destination node $e_o^t$ with type $r$.

Extrapolated TKG reasoning is the task of predicting the evolution of KGs over time, and we perform link prediction at future times. It is also forecasting of events occurring in the near future. Given a query $(e_q, r_q, ?, t_q)$ or $(?, r_q, e_q, t_q)$, we have a set of known facts $\{(e_{s_i}, r_i, e_{o_i}, t_i)|t_i < t_q\}$.
These known facts constitute the known TKG,
and our goal is to predict the missing object or subject entity in the query. 

\subsection{Reinforcement Learning Framework} \label{environment}
\begin{figure}[]
  \centering
  \includegraphics[width=\linewidth]{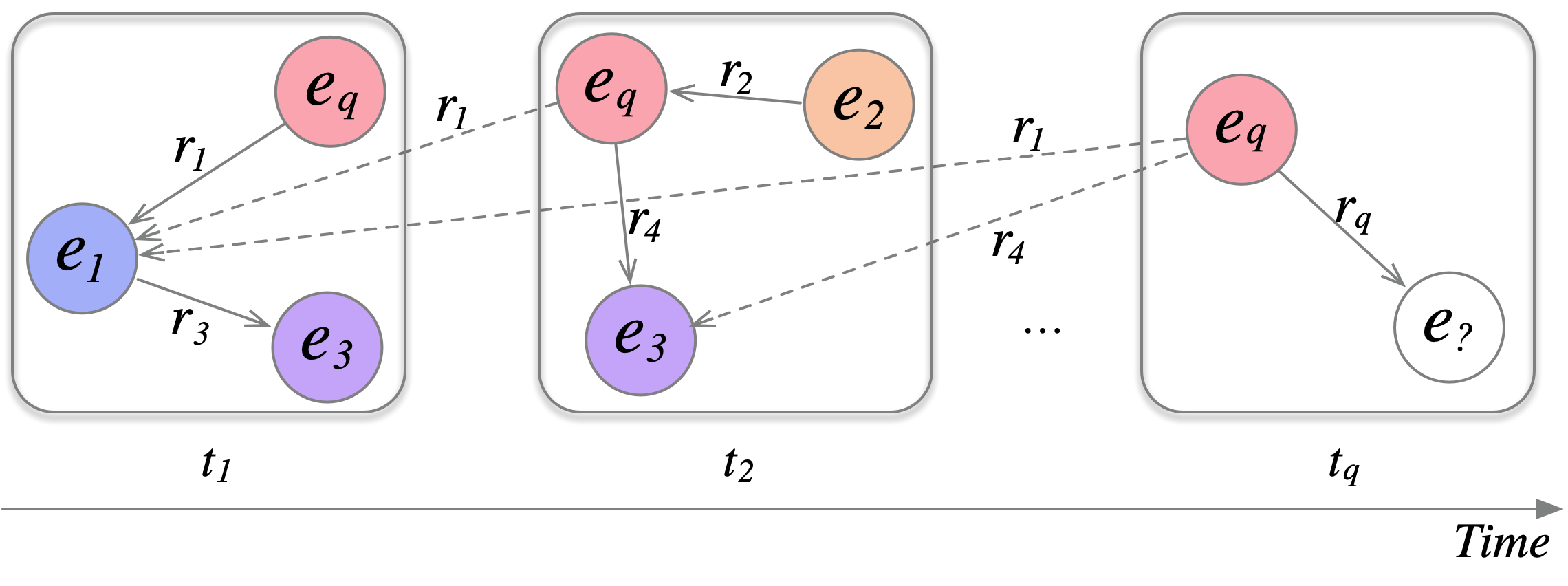}
  \caption{
Illustration of the TKG with temporal edges. To ensure the figure be clear enough, we omit the self-loop edges and reversed edges. The dotted lines are temporal edges.}
  \label{graph}
\end{figure}

Because there is no edge among the typical TKG snapshots, the agent cannot transfer from one snapshot to another. Hence, we sequentially add three types of edges. 
(\romannumeral1) \textit{Reversed Edges}. For each quadruple $(e_s, r, e_o, t)$, we add $(e_o, r^{-1}, e_s, t)$ to the TKG, where $r^{-1}$ indicates the reciprocal relation of $r$. Thus, we can predict the subject entity by converting $(?, r, e_q, t)$ to $(e_q, r^{-1}, ?, t)$ without loss of generality.   
(\romannumeral2) \textit{Self-loop Edges}. Self-loop edges can allow the agent to stay in a place and work as a stop action when the agent search unrolled for a fixed number of steps. 
(\romannumeral3) \textit{Temporal Edges}. The agent can walk from node $e_s^{t_j}$ to node $e_o^{t_i}$ through edge $r$, if $(e_s, r, e_o, t_i)$ exits and $t_i < t_j \leq t_q$. Temporal edges indicate the impact of the past fact on the entity and help the agent find the answer in historical facts. Figure \ref{graph} shows the graph with temporal edges. 


  

Our method can be formulated as a Markov Decision Process (MDP), and the components of which are elaborated as follows.

\paragraph{States.} Let $\mathcal{S}$ denote the state space, 
and a state is represented by a quintuple $s_l = (e_{l}, t_{l}, e_q, t_q, r_q) \in \mathcal{S}$, where $(e_l, t_l)$ is the node visited at step $l$ and $(e_q, t_q, r_q)$ is the elements in the query. $(e_q, t_q, r_q)$ can be viewed as the global information while $(e_l, t_l)$ is the local information. The agent starts from the source node of the query, so the initial state is $s_0 = (e_q, t_q, e_q, t_q, r_q)$.

\paragraph{Actions.} Let $\mathcal{A}$ denote the action space, and $\mathcal{A}_l$ 
denote the set of optional actions at step $l$, $\mathcal{A}_l \subset \mathcal{A}$ consists of outgoing edges of node $e_l^{t_l}$. Concretely, $\mathcal{A}_l$ should be $\{(r', e', t')|(e_l, r', e', t') \in \mathcal{F}, t' \leq t_l, t' < t_q \}$, but an entity usually has many related historical facts, leading to a large number of optional actions. Thus, the final set of optional actions $\mathcal{A}_l$ is sampled from the set of above outgoing edges.

\paragraph{Transition.} The environment state is transferred to a new node through the edge selected by the agent. The transition function $\delta: \mathcal{S} \times \mathcal{A} \rightarrow \mathcal{S}$ defined by $\delta(s_l, \mathcal{A}_l) = s_{l+1} = (e_{l+1}, t_{l+1}, e_q, t_q, r_q)$, where $\mathcal{A}_l$ is the sampled outgoing edges of $e_l^{t_l}$.

\paragraph{Rewards with shaping.} The agent receives a terminal reward of 1 if it arrives at a correct target entity at the end of  the search and 0 otherwise. If $s_L = (e_L, t_L, e_q, t_q, r_q)$ is the final state and $(e_q, r_q, e_{gt}, t_q)$ is the ground truth fact, the reward formulation is: 
\begin{equation} \label{reward}
    R(s_L) = \mathbb{\uppercase\expandafter{\romannumeral1}}\{e_L == e_{gt}\}.
\end{equation}
Usually, the quadruples with the same entity are concentrated in specific periods, which causes temporal variability and temporal sparsity~\cite{DBLP:conf/emnlp/WuCCH20}. Due to such property, the answer entity of the query has a distribution over time, and we can introduce this prior knowledge into the reward function to guide the agent learning. The time-shaped reward can let the agent know 
which snapshot is more likely to find the answer. 
Based on the training set, we estimate a Dirichlet distribution for each relation. Then, we shape the original reward with Dirichlet distributions:
\begin{equation}
\begin{split}
    \widetilde{R}(s_L) &= (1 + p_{\Delta t_L}) R(s_L), \\
    \Delta t_L &= t_q - t_L, \\
    (p_1,..., p_K) &\sim Dirichlet(\bm{\alpha}_{r_q}),
\end{split}
\end{equation}
where $\bm{\alpha}_{r_q} \in \mathbb{R}^K$ is a vector of parameters of the Dirichlet distribution for a relation $r_q$. We can estimate $\bm{\alpha}_{r_q}$ from the training set. For each quadruple with relation $r_q$ in the training set, we count the number of times the object entity appears in each of the most recent $K$ historical snapshots. Then, we obtain a multinomial sample $x_i$ and $D = \{x_1, ..., x_N\}$. To maximize the likelihood:
\begin{equation}
     p(D|\bm{\alpha}_{r_q}) = \prod_i p(x_i|\bm{\alpha}_{r_q}),
\end{equation}
we can estimate $\bm{\alpha}_{r_q}$. The maximum can be computed via the fixed-point iteration, and more calculation formulas are provided in Appendix \ref{estimating a Dirichlet distribution}. Because Dirichlet distribution has a conjugate prior, we can update it easily when we have more observed facts to train the model.

\subsection{Policy Network} \label{policy}
We design a policy network $\pi_{\theta}(a_l|s_l)=P(a_l|$ $s_l;\theta)$ to model the agent in a continuous space, where $a_l \in \mathcal{A}_l$, and $\theta$ are the model parameters. The policy network consists of the following three modules.

\paragraph{Dynamic embedding.} We assign each relation $r \in \mathcal{R}$ a dense vector embedding $\mathbf{r} \in \mathbb{R}^{d_r}$. As the characteristic of entities may change over time, we adopt a relative time representation method for entities. We use a dynamic embedding to represent each node $e^t_i = (e_i, t)$ in $\mathcal{G}_t$, and use $\mathbf{e} \in \mathbb{R}^{d_e}$ to represent the latent invariant features of entities. We then define a relative time encoding function $\boldsymbol{\Phi}(\Delta t) \in \mathbb{R}^{d_t}$ to represent the time information.  $\Delta t = t_q - t$ and $\boldsymbol{\Phi}(\Delta t)$ is formulated as follows:
\begin{equation}
    \boldsymbol{\Phi}(\Delta t) = \sigma(\mathbf{w}\Delta t + \mathbf{b}),
\end{equation}
where $\mathbf{w}, \mathbf{b} \in \mathbb{R}^{d_t}$ are vectors with learnable parameters and $\sigma$ is an activation function.
$d_r$, $d_e$ and $d_t$ represent the dimensions of the embedding. Then, we can get the representation of a node $e^t_i$: $\mathbf{e}^t_i = [\mathbf{e}_i;\boldsymbol{\Phi}(\Delta t)]$. 

\paragraph{Path encoding.} The search history $h_l = ((e_q, t_q), r_1, (e_1, t_1),..., r_l, (e_l, t_l))$ is the sequence of actions taken. The agent encodes the history $h_l$ with a LSTM:
\begin{equation}
    \begin{aligned}
    \mathbf{h}_l &= {\rm LSTM}(\mathbf{h}_{l-1}, [\mathbf{r}_{l-1};\mathbf{e}_{l-1}^{t_{l-1}}]), \\
    \mathbf{h}_0 &= {\rm LSTM}(\mathbf{0}, [\mathbf{r}_0;\mathbf{e}_q^{t_q}]).
    \end{aligned}
\end{equation}
Here, $\mathbf{r}_0$ is a start relation, and we keep the LSTM state unchanged when the last action is self-loop.

\paragraph{Action scoring.} We score each optional action and calculate the probability of state transition. Let $a_n = (e_n, t_n, r_n) \in \mathcal{A}_l$ represent an optional action at step $l$. Future events are usually uncertain, and there is usually no strong causal logic chain for some queries, so the correlation between the entity and query is sometimes more important. Thus, we use a weighted action scoring mechanism to help the agent pay more attention to attributes of the destination nodes or types of edges. 
Two Multi-Layer Perceptrons (MLPs) are used to encode the state information and output expected destination node $\widetilde{e}$ and outgoing edge $\widetilde{r}$ representations. Then, the agent obtains the destination node score and outgoing edge score of the candidate action by calculating the similarity. With the weighted sum of the two scores, the agent obtains the final candidate action score $\phi(a_n, s_l)$:
\begin{equation}
\label{attention}
    \phi(a_n, s_l) = \beta_n \Braket{\widetilde{\mathbf{e}}, \mathbf{e}_n^{t_n}} + (1 - \beta_n) \Braket{\widetilde{\mathbf{r}}, \mathbf{r}_n}, \\
\end{equation}
\begin{shrinkeq}{-3.3ex}
\begin{equation}
\begin{split}
    \widetilde{\mathbf{e}} = \mathbf{W}_e {\rm ReLU}( \mathbf{W}_1[\mathbf{h}_l; \mathbf{e}_q^{t_q}; \mathbf{r}_q]), \\
    \widetilde{\mathbf{r}} = \mathbf{W}_r {\rm ReLU}( \mathbf{W}_1[\mathbf{h}_l; \mathbf{e}_q^{t_q}; \mathbf{r}_q]), \\
\end{split}
\end{equation}
\end{shrinkeq}
\begin{equation}
    \beta_n = {\rm sigmoid}(\mathbf{W}_\beta[\mathbf{h}_l; \mathbf{e}_q^{t_q}; \mathbf{r}_q; \mathbf{e}_n^{t_n}; \mathbf{r}_n]),
\end{equation}
where $\mathbf{W}_1$, $\mathbf{W}_e$, $\mathbf{W}_r$ and $\mathbf{W}_{\beta}$ are learnable matrices. After scoring all candidate actions in $\mathcal{A}_l$, $\pi_{\theta}(a_l|s_l)$ can be obtained through softmax.

To summarize, the parameters of the LSTM, MLP and $\boldsymbol{\Phi}$, the embedding matrices of relation and entity form the parameters in $\theta$.

\subsection{Optimization and Training} \label{training}
We fix the search path length to $L$, and an $L$-length trajectory will be generated from the policy network $\pi_{\theta}$: $\{a_1, a_2, ..., a_L\}$. The policy network is trained by maximizing the expected reward over all training samples $\mathcal{F}_{train}$: 

\begin{equation}
\begin{split}
    J(\theta) = \mathbb{E}_{(e_s, r, e_o, t) \sim \mathcal{F}_{train}}[\mathbb{E}_{a_1,...,a_L \sim \pi_{\theta}} \\ [\widetilde{R}(s_L|e_s, r, t)]].
\end{split}
\end{equation}

Then, we use the policy gradient method to optimize the policy. The REINFORCE algorithm~\cite{williams1992simple} will iterate through all quadruple in $\mathcal{F}_{train}$ and update $\theta$ with the following stochastic gradient:
\begin{equation}
    \nabla_\theta J(\theta) \approx \nabla_\theta \sum_{m \in [1, L]} \widetilde{R}(s_L|e_s, r, t)log \pi_\theta(a_l|s_l)
\end{equation}


\begin{figure}[]
  \centering
  \includegraphics[width=0.9\linewidth]{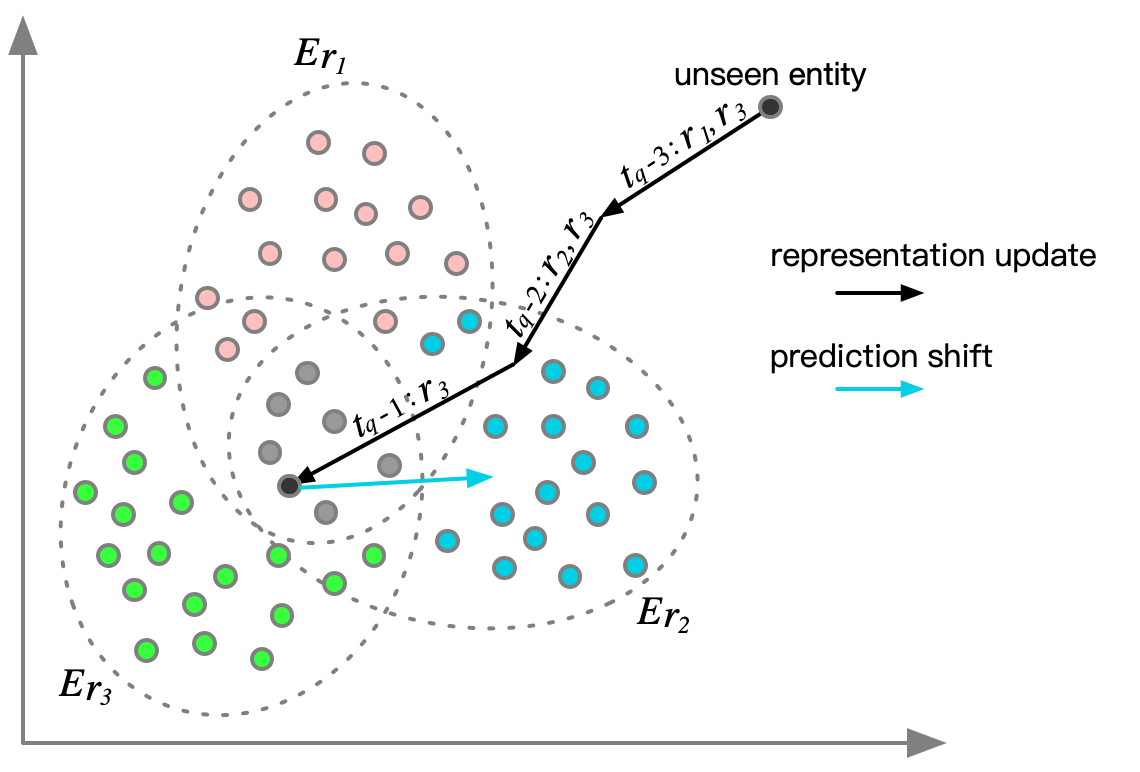}
  \caption{Illustration of the IM mechanism. 
  For an unseen entity $e_q$, ``$t_q-3: r_1, r_3$'' indicates $e_q$ has co-occurrence relations $r_1, r_3$ at $t_q-3$, and updates its representation based on $E_{r_1}, E_{r_3}$, and finally gets the IM representation at $t_q-1$. Then to answer a query $(e_{q},r_2,?,t_q)$, we do a prediction shift based on 
$E_{r_2}$.
  }
  \vspace{-0.5em}
  \label{Figure:IM}
\end{figure}

\subsection{Inductive Mean Representation} 
\label{inference}
As new entities always emerge over time, we propose a new entity representation method for previously unseen entities.
Previous works~\cite{bhowmik2020explainable,han2021explainable} can represent unseen entities through neighbor information aggregation. However, newly emerging entities usually have very few links, which means that only limited information is available. E.g., for a query $(Evan\_Mobley,plays\_for,?,2022)$, entity ``Evan\_Mobley'' does not exist in previous times, but we could infer this entity is a player through relation ``plays\_for'', and assign ``Evan\_Mobley'' a more reasonable initial embedding that facilitates the inference. Here we provide another approach to represent unseen entities by leveraging the query information and embeddings of the trained entities, named Inductive Mean (IM), as illustrated in Figure \ref{Figure:IM}.

Let $\mathcal{G}_{(t_j, t_q-1)}$ represent the snapshots of the TKG in the test set. The query entity $e_q$ first appears in $\mathcal{G}_{t_j}$ and gets a randomly initialized representation vector.
We regard $r$ as the \textit{co-occurrence relation} of $e_q$, if there exists a quadruple which contains $(e_q, r)$.
Note that entity $e_q$ may have different co-occurrence relations over time.
We denote $R_t(e_q)$ as the co-occurrence relation set of entity $e_q$ at time $t$. 
Let $E_r$ represent the set that comprises all the trained entities having the co-occurrence relation $r$. 
Then, we can obtain the inductive mean representation of the entities with the same co-occurrence relation $r$: 
\begin{equation}
    \overline{\mathbf{e}^{r}} = \frac{\sum_{e \in E_r} \mathbf{e}}{|E_r|}.
\end{equation}

Entities with the same co-occurrence relation $r$ have similar characteristics, so IM can utilize $\overline{\mathbf{e}^{r}}$ to gradually update the representation of $e_q$ based on the time flow. $0 \leq \mu \leq 1$ is a hyperparameter:
\begin{equation}
    \mathbf{e}_{q, t} = \mu\mathbf{e}_{q, t-1} + (1 - \mu)\frac{\sum_{r \in R_t(e_q)} \overline{\mathbf{e}^{r}}}{|R_t(e_q)|}.
\end{equation}

For relation $r_q$, we do a prediction shift based on  $\overline{\mathbf{e}^{r_q}}$ to make the entity representation more suitable for the current query. To answer a query $(e_q, r_q, ?, t_q)$, we use the combination of $e_q$'s representation at time $t_q-1$ and the inductive mean representation $\overline{\mathbf{e}^{r_q}}$:

\begin{equation}
    \mathbf{e}_{q, t_q, r_q} = \mu\mathbf{e}_{q, t_q-1} + (1 - \mu) \overline{\mathbf{e}^{r_q}}.
\end{equation}

\begin{table*}[]
\centering
\renewcommand{\arraystretch}{1.3}
\caption{Comparison on future link prediction. 
The results of MRR and Hits@1/3/10 are multiplied by 100. The best results are in bold. We average the metrics over five runs.}
\label{table:experiment_results_tdf}
\resizebox{\linewidth}{!}{
\begin{tabular}{lcccccccccccccccc}
\hline\hline
\multicolumn{1}{c}{\multirow{2}{*}{\textbf{Method}}} & \multicolumn{4}{c}{\textbf{ICEWS14}}                               & \multicolumn{4}{c}{\textbf{ICEWS18}}                                        & \multicolumn{4}{c}{\textbf{WIKI}}                                           & \multicolumn{4}{c}{\textbf{YAGO}}                      \\ \cline{2-17} 
\multicolumn{1}{c}{}                                 & MRR   & H@1          & H@3          & H@10                              & MRR            & H@1          & H@3          & H@10                              & MRR            & H@1          & H@3          & H@10                              & MRR            & H@1          & H@3          & H@10         \\ \hline
\multicolumn{1}{l|}{TTransE}                         & 13.43 & 3.11           & 17.32          & \multicolumn{1}{c|}{34.55}          & 8.31           & 1.92           & 8.56           & \multicolumn{1}{c|}{21.89}          & 29.27          & 21.67          & 34.43          & \multicolumn{1}{c|}{42.39}          & 31.19          & 18.12          & 40.91          & 51.21          \\
\multicolumn{1}{l|}{TA-DistMult}                     & 26.47 & 17.09          & 30.22          & \multicolumn{1}{c|}{45.41}          & 16.75          & 8.61           & 18.41          & \multicolumn{1}{c|}{33.59}          & 44.53          & 39.92          & 48.73          & \multicolumn{1}{c|}{51.71}          & 54.92          & 48.15          & 59.61          & 66.71          \\
\multicolumn{1}{l|}{DE-SimplE}                       & 32.67 & 24.43          & 35.69          & \multicolumn{1}{c|}{49.11}          & 19.30          & 11.53          & 21.86          & \multicolumn{1}{c|}{34.80}          & 45.43          & 42.6           & 47.71          & \multicolumn{1}{c|}{49.55}          & 54.91          & 51.64          & 57.30          & 60.17          \\
\multicolumn{1}{l|}{TNTComplEx}                      & 32.12 & 23.35          & 36.03          & \multicolumn{1}{c|}{49.13}          & 27.54          & 19.52          & 30.80          & \multicolumn{1}{c|}{42.86}          & 45.03          & 40.04          & 49.31          & \multicolumn{1}{c|}{52.03}          & 57.98          & 52.92          & 61.33          & 66.69          \\ \hline
\multicolumn{1}{l|}{CyGNet}                          & 32.73 & 23.69          & 36.31          & \multicolumn{1}{c|}{50.67}          & 24.93          & 15.90          & 28.28          & \multicolumn{1}{c|}{42.61}          & 33.89          & 29.06          & 36.10          & \multicolumn{1}{c|}{41.86}          & 52.07          & 45.36          & 56.12          & 63.77          \\
\multicolumn{1}{l|}{RE-NET}                          & 38.28 & 28.68          & 41.34          & \multicolumn{1}{c|}{54.52}          & 28.81          & 19.05          & 32.44          & \multicolumn{1}{c|}{\textbf{47.51}}          & 49.66          & 46.88          & 51.19          & \multicolumn{1}{c|}{53.48}          & 58.02          & 53.06          & 61.08          & 66.29          \\
\multicolumn{1}{l|}{xERTE}                           & 40.79 & 32.70 & 45.67          & \multicolumn{1}{c|}{57.30}          & 29.31          & 21.03          & \textbf{33.51} & \multicolumn{1}{c|}{46.48} & 71.14          & 68.05          & 76.11          & \multicolumn{1}{c|}{79.01}          & 84.19          & 80.09          & 88.02          & 89.78          \\
\multicolumn{1}{l|}{TANGO-Tucker}                     & \---     & \---              & \---              & \multicolumn{1}{c|}{\---}              & 28.68          & 19.35          & 32.17          & \multicolumn{1}{c|}{47.04}          & 50.43          & 48.52          & 51.47          & \multicolumn{1}{c|}{53.58}          & 57.83          & 53.05          & 60.78          & 65.85          \\
\multicolumn{1}{l|}{TANGO-DistMult}                   & \---     & \---              & \---              & \multicolumn{1}{c|}{\---}              & 26.75          & 17.92          & 30.08          & \multicolumn{1}{c|}{44.09}          & 51.15          & 49.66          & 52.16          & \multicolumn{1}{c|}{53.35}          & 62.70          & 59.18          & 60.31          & 67.90          \\ \hline
\multicolumn{1}{l|}{TITer}                           & \textbf{41.73} & \textbf{32.74}          & \textbf{46.46} & \multicolumn{1}{c|}{\textbf{58.44}} & \textbf{29.98} & \textbf{22.05} & 33.46          & \multicolumn{1}{c|}{44.83}          & \textbf{75.50} & \textbf{72.96} & \textbf{77.49} & \multicolumn{1}{c|}{\textbf{79.02}} & \textbf{87.47} & \textbf{84.89} & \textbf{89.96} & \textbf{90.27} \\ \hline\hline
\end{tabular}
}
\end{table*}

\section{Experiments}
\label{sec:exp}

\subsection{Experimental Setup}
\paragraph{Datasets.} We use four public TKG datasets for evaluation: ICEWS14, ICEWS18~\cite{DVN/28075_2015}, WIKI~\cite{LeblayDeriving2018}, and YAGO~\cite{MahdisoltaniBS15}. 
Integrated Crisis Early Warning System (ICEWS) is an event dataset.
ICEWS14 and ICEWS18 are two subsets of events in ICEWS that occurred in 2014 and 2018 with a time granularity of days.
WIKI and YAGO are two knowledge base that contains facts with time information, and we use the subsets with a time granularity of years.
We adopt the same dataset split strategy as in~\cite{jin2020Renet} and split the dataset into train/valid/test by timestamps such that:\\
$~~~~\text{time\_of\_train} < \text{time\_of\_valid} < \text{time\_of\_test}$.
Appendix \ref{Dataset Statistics} summarizes more statistics on the datasets.

\paragraph{Evaluation metrics.} We evaluate our model on TKG forecasting, a link prediction task at the future timestamps. Mean Reciprocal Rank (MRR) and Hits@1/3/10 are performance metrics. For each quadruple $(e_s, r, e_o, t)$ in the test set, 
we evaluate two queries, $(e_s, r, ?, t)$ and $(?, r, e_o, t)$. We use the time-aware filtering scheme ~\cite{han2020graph} that only filters out quadruples with query time $t$. The time-aware scheme is more reasonable than the filtering scheme used in~\cite{jin2020Renet, zhu-etal-2021-cygnet}. 
Appendix \ref{Definitions for Evaluation Metrics} provides detailed definitions.

\paragraph{Baseline.}
As lots of previous works have verified that the static methods underperform compared with the temporal methods on this task,  
we do not compare TITer with them. 
We compare our model with existing interpolated TKG reasoning methods, including TTransE~\cite{leblay2018deriving}, TA-DistMult~\cite{DBLP:conf/emnlp/Garcia-DuranDN18}, DE-SimplE~\cite{DBLP:conf/aaai/GoelKBP20}, and TNTComplEx~\cite{DBLP:conf/iclr/LacroixOU20}, and state-of-the-art extrapolated TKG reasoning approaches, including RE-NET~\cite{jin2020Renet}, CyGNet~\cite{zhu-etal-2021-cygnet}, TANGO~\cite{ding2021temporal}, and xERTE~\cite{han2021explainable}. An overview of these methods is in Section \ref{sec:relatedWork}.


\subsection{Implementation Details}
Our model is implemented in PyTorch\footnote{https://github.com/JHL-HUST/TITer/}. We set the entity embedding dimension to 80, the relation embedding dimension to 100, 
and the relative time encoding dimension to 20. 
We choose the latest $N$ outgoing edges as candidate actions for TITer at each step.
$N$ is 50 for ICEWS14 and ICEWS18, 60 for WIKI, and 30 for YAGO. 
The reasoning path length is 3. The discount factor $\gamma$ of REINFORCE is 0.95. We use Adam optimizer to optimize the parameters, and the learning rate is 0.001. 
The batch size is set to 512 during training. We use beam search for inference, and the beam size is 100. For the IM, $\mu$ is 0.1. The activation function of $\boldsymbol{\Phi}$ is $cosine$. For full details, please refer to Appendix \ref{Detailed Implementation}.

\begin{table}[]\footnotesize
\centering
\renewcommand{\arraystretch}{1.4}
\caption{The percentage of quaduples containing unseen entities of used test datasets.}
\label{table:unseen_entities_rate}
\begin{tabular}{lcccc}
\hline\hline
\multicolumn{1}{c}{\textbf{Datasets}} & \textbf{ICEWS14} & \textbf{ICEWS18} & \textbf{WIKI} & \textbf{YAGO} \\ \hline
Proportion                   & 6.52    & 3.93    & 42.91 & 8.03 \\ \hline\hline
\end{tabular}
\end{table}

\begin{table}[]\footnotesize
\centering
\renewcommand{\arraystretch}{1.2}
\caption{Comparison on the number of parameters and calculation. (M means million.)
}
\label{table:params_and_macs}
\begin{tabular}{lcr}
\hline\hline
\multicolumn{1}{c}{Method} & \# Params & \# MACs     \\ \hline
RE-NET                     & 5.459M & 4.370M    \\
CyGNet                     & 8.568M & 8.554M   \\
xERTE(3 steps) 
& 2.927M & 225.895M \\
TITer(3 steps)              & 1.455M & 0.225M     \\ \hline\hline
\end{tabular}
\end{table}

\subsection{Results and Discussion}

\paragraph{Performance on the TKG datasets.} Our method can search multiple candidate answers via beam search. Table \ref{table:experiment_results_tdf} reports the TKG forecasting performance of TITer and the baselines on four TKG datasets. TITer outperforms all baselines on all datasets when evaluated by MRR and Hits@1 metrics, and in most cases (except ICEWS18),
TITer exhibits the best performance when evaluated by the other two metrics. 
TTransE, TA-DistMult, DE-SimplE, and TNTComplEx cannot deal with unseen timestamps in the test set, so they perform worse than others. 
The performance of TITer is much higher than RE-NET, CyGNet, and TANGO on WIKI and YAGO. 
Two reasons cause this phenomenon: (1) 
the characteristic of WIKI and YAGO that nodes usually have a small number of neighbors gives the neighbor search algorithm an advantage.
(2) for WIKI and YAGO, a large number of quadruples contain unseen entities in the test set(See Table \ref{table:unseen_entities_rate}), but these methods cannot handle these queries.

\begin{figure}
    \centering
    \includegraphics[width=\linewidth]{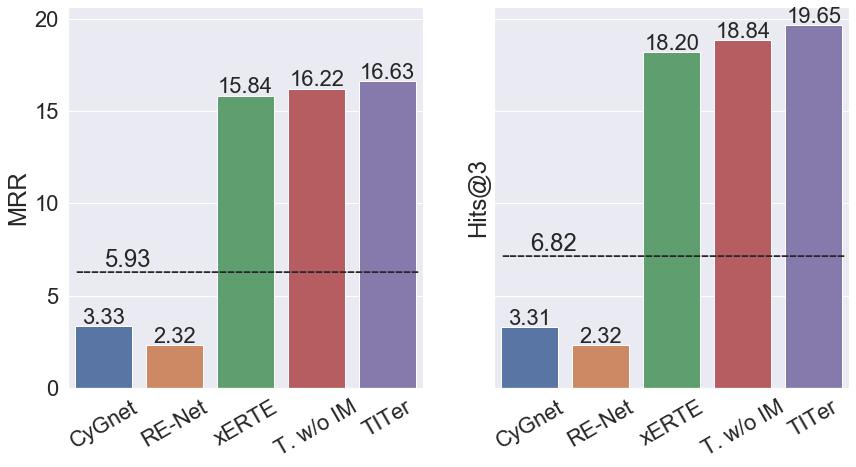}
    \caption{Inductive inference results on a subset of ICEWS14 that contain unseen entities. The dotted line corresponds to the score of a 3-steps random search. T. w/o IM: TITer without IM mechanism.}
    \label{figure:inductive_mrr_hit10}
\end{figure}

\paragraph{Inductive inference.} 
When a query contains an unseen entity, models should infer the answer inductively. For all such queries that contain unseen entities in the ICEWS14 test set, we present experimental results in Figure \ref{figure:inductive_mrr_hit10} and Table \ref{table:ablation_IM_ICE14_WIKI}. 
The performance of RE-NET and CyGNet 
decays significantly when compared with their result on the whole ICEWS14 test set (see Table \ref{table:experiment_results_tdf}). 
Due to the lack of training for unseen entities' embedding and the classification layer for all entities, RE-Net and CyGNet could not reach the performance of a 3-hop neighborhood random search baseline, as illustrated by the dotted line in Figure \ref{figure:inductive_mrr_hit10}. 
In contrast, xERTE can tackle such queries by dynamically updating the new entities' representation based on 
temporal message aggregation, 
and TITer can tackle such queries by the temporal-path-based RL framework.
We also observe that TITer outperforms xERTE, no matter TITer adopts the IM mechanism or not.
The IM mechanism could further boost the performance,
demonstrating its effectiveness in representing unseen entities.

\paragraph{Case study.} 
Table \ref{Visualization} visualizes specific reasoning paths of several examples in the test set of ICEWS18. 
We notice that TITer tends to select a recent fact (outgoing edge) to search for the answer.
Although the first two queries have the same subject entity and relation, TITer can reach the ground truths according to different timestamps. 
As shown in Eq. (\ref{attention}),  $\beta$ increases when TITer emphasizes neighbor nodes more than edges. After training, the representations of entities accumulate much semantic information, which helps TITer select the answer directly with less extra information for queries 3 and 4.
In comparison, TITer needs more historical information when unseen entities appear. Query 5 is an example of multi-hop reasoning. It indicates that TITer can tackle combinational logic problems.

\begin{table*}[]\scriptsize
\caption{Path visualization on ICEWS18. $\dag$ indicates unseen entities not appeared in the training set. $\ddag$ indicates entities along a path with the same country. $\beta$ is defined in Eq. (\ref{attention}). For a test quadruple, we use the object prediction as an example.}
\label{Visualization}
\centering
\renewcommand{\arraystretch}{2}
\begin{tabular}{@{}m{\tabcolsep}@{}m{4pt}<{\centering}|m{148pt}|m{245pt}|m{10pt}<{\centering}@{}m{\tabcolsep}@{}}
\hline
\hline
\rule[0pt]{0pt}{10pt} &
\textbf{ID} &
  \textbf{Test quadruple} &
  \textbf{Path} &
  \multicolumn{1}{l}{$\beta$} &
\rule[0pt]{0pt}{1pt}
\\ \hline
\rule[0pt]{0pt}{20pt} &
1 &
  (Military (Comoros)$\dag$, Use conventional military force, Citizen (Comoros), 2018/10/17) &
  Military$\ddag$$\xrightarrow[t=2018/10/16]{fight\;with\;small\;arms\;and\;light\;weapons}$Citizen$\ddag$ &
  0.46 &
\rule[0pt]{0pt}{1pt}
\\ \hline
\rule[0pt]{0pt}{20pt} &
2 &
  (Military (Comoros)$\dag$, Use conventional military force, Armed Rebel (Comoros)$\dag$, 2018/10/26) &
  Military$\ddag$ $\xrightarrow[t=2018/10/19]{Investigate^{-1}}$Armed Rebel$\ddag$ &
  0.54 &
\rule[0pt]{0pt}{1pt}
\\ \hline
\rule[0pt]{0pt}{20pt} &
3 &
  (Nigeria, Consult, Muhammadu Buhari, 2018/10/04) &
  Nigeria$\xrightarrow[t=2018/10/01]{Make\;an\;appeal\;or\;request^{-1}}$Muhammadu Buhari &
  0.96 &
\rule[0pt]{0pt}{1pt}
\\ \hline
\rule[0pt]{0pt}{20pt} &
4 &
  (Police (India), Physically assault, Citizen (India), 2018/10/08) &
  Police$\ddag$$\xrightarrow[t=2018/10/06]{Investigate}$Citizen$\ddag$ &
  0.96 &  
\rule[0pt]{0pt}{1pt}
\\ \hline
\rule[0pt]{0pt}{20pt} &
5 &
  (Governor (Cote d'Ivoire)$\dag$, Make an appeal or request, Citizen (Cote d'Ivoire), 2018/10/14) &
  Governor$\ddag$$\xrightarrow[t=2018/10/12]{Praise\;or\;endorse}$Party Member$\ddag$$\xrightarrow[t=2018/9/29]{Make\;an\;appeal\;or\;request}$Citizen$\ddag$ &
  \--- &
\rule[0pt]{0pt}{1pt}
\\ \hline
\hline
\end{tabular}
\end{table*}

\paragraph{Efficiency analysis.} 
Table \ref{table:params_and_macs} reflects the complexity of RE-NET, CyGNet, xERTE, and TITer. Due to the enormous linear classification layers, RE-NET and CyGNet have much more parameters than other methods. To achieve the best results, xERTE adopts the graph expansion mechanism and the temporal relational graph attention layer to perform a local representation aggregation for each step, leading to a vast amount of calculation. 
Compared with xERTE, the number of parameters of TITer has reduced by at least a half, and the number of Multi-Adds operations (MACs) has greatly reduced to 0.225M, which is much less than the counterpart, indicating the high efficiency of the proposed model. 

In summary, compared to the previous state-of-the-art models, TITer has saved at least 50.3\% parameters and 94.9\% MACs. Meanwhile, TITer still exhibits better performance.

\begin{table}[]
\renewcommand{\arraystretch}{1.2}
\centering
\caption{Ablation study on ICEWS18. w/o: without, ws: weighted action scoring mechanism, rs: reward shaping.}
\label{table:ablation_study}
\resizebox{\linewidth}{!}{
\begin{tabular}{lcccc}
\hline\hline
\multicolumn{1}{c}{Method}   & MRR            & H@1            & H@3            & H@10           \\ \hline
TITer w/o ws and rs   & 29.13          & 20.73          & 32.74          & 45.04          \\
TITer w/o ws    & 29.25          & 20.84          & 32.83          & \textbf{45.06}          \\
TITer w/o rs & 29.17          & 21.00          & 32.72          & 44.51 \\ \hline
TITer                        & \textbf{29.98} & \textbf{22.05} & \textbf{33.46} & 44.83          \\ \hline\hline
\end{tabular}
}
\end{table}

\begin{table}[]\footnotesize
\caption{Results improvement with IM mechanism on subsets of ICEWS14 and WIKI that contain unseen entities.}
\label{table:ablation_IM_ICE14_WIKI}
\renewcommand{\arraystretch}{1.3}
\centering
\begin{tabular}{lcccc}
\hline\hline
\multicolumn{1}{c}{Datasets} & MRR   & H@1   & H@3   & H@10  \\ \hline
ICEWS14                      & +0.41 & +0.46 & +0.81 & +0.01 \\
WIKI                         & +1.52 & +1.15 & +2.15 & +1.90 \\ \hline\hline
\end{tabular}
\end{table}

\subsection{Ablation Study} 
\begin{figure}
    \centering
    \includegraphics[width=\linewidth]{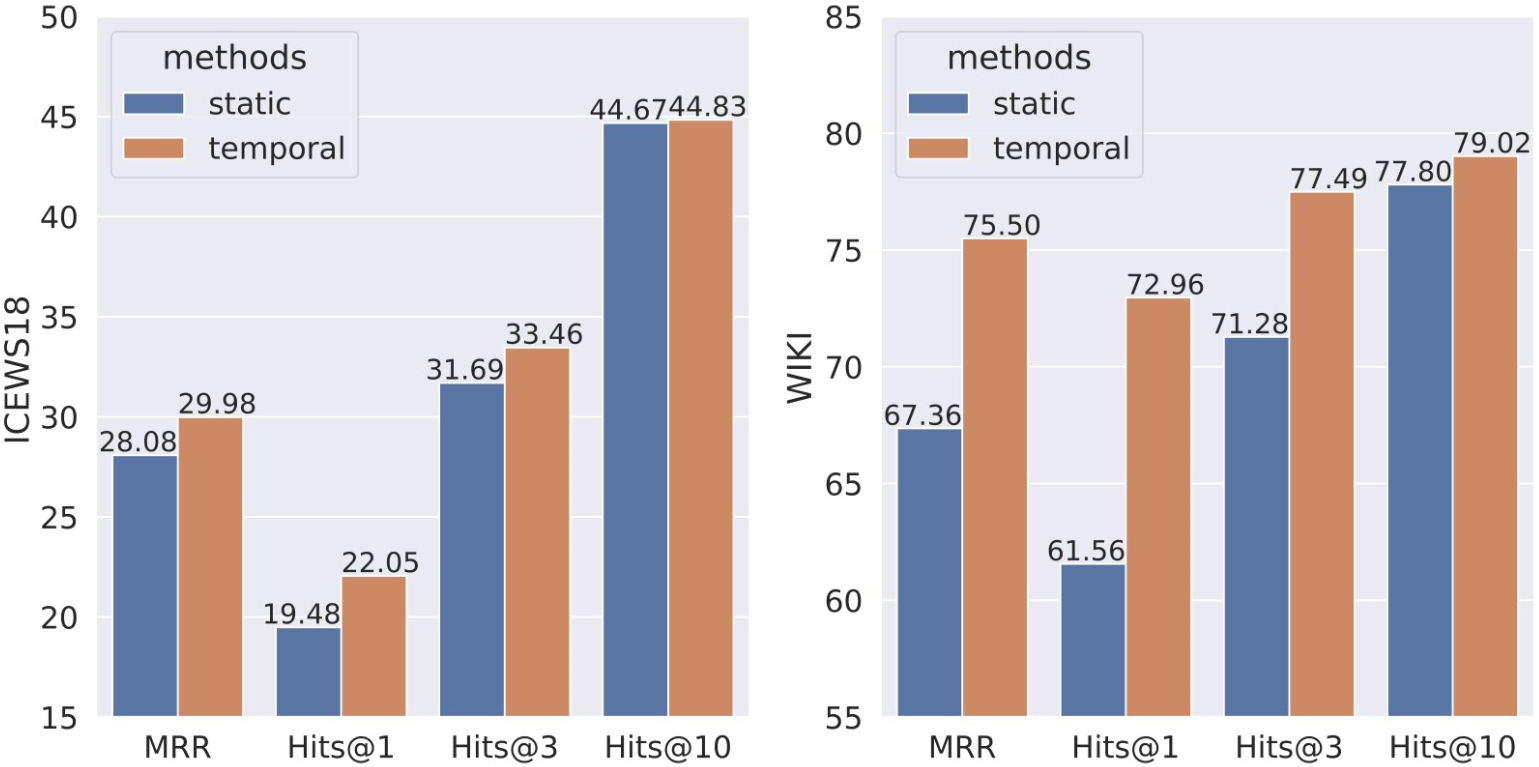}
    \caption{Ablation of time information.}
    \label{figure:dynamic embedding}
\end{figure}

In this subsection, we study the effect of different components of TITer by 
ablation studies.
The results are shown in Table \ref{table:ablation_study} and Figure \ref{figure:dynamic embedding}. 

\paragraph{Relative time encoding.}
The relative time representation is a crucial component in our method. Figure \ref{figure:dynamic embedding} shows the notable change from temporal to static on ICEWS18 and WIKI. We remove the relative time encoding module to get the static model. For Hits@1, the temporal model improves 13.19\% on ICEWS18 and 18.51\% on WIKI, compared with the static model. It indicates that our relative time encoding function can 
help TITer choose the correct answer more precisely.

\paragraph{Weighted action scoring mechanism.} We observe that setting $\beta$ to a constant 0.5 can lead to a drop of 5.49\% on Hits@1, indicating that TITer can better choose the source of evidence when making the inference. 
After training, TITer learns the latent relationship among entities. As expounded in Table \ref{Visualization}, TITer prefers to pay more attention to the node for inferring when there exists more information to make a decision, and TITer chooses to focus on edges (relations in history) to assist the inferring for complex queries or unseen entities.

\paragraph{Reward shaping.} We observe that TITer outperforms the variant without reward shaping, which means an improvement of 5\% on Hits@1. By using Dirichlet prior distribution to direct the decision process, TITer acquires knowledge about the probability distribution of the target's appearance over the whole time span.

\section{Conclusion}
In this work, we propose a temporal-path-based reinforcement learning model named TimeTraveler (TITer) for 
temporal knowledge graph forecasting. TITer travels on the TKG historical snapshots and searches for the temporal evidence chain to find the answer. TITer uses a relative time encoding function and time-shaped reward to model the time information, and the IM mechanism to update the unseen entities' representation in the process of testing.
Extensive experimental results reveal that our model 
outperforms state-of-the-art baselines 
with less calculation and fewer parameters. Furthermore, the inference process of TITer is explainable, and TITer has good inductive reasoning ability. 

\section*{Acknowledgements}
This work is supported by National Natural Science Foundation (62076105). 

\bibliography{emnlp2021}

\begin{thebibliography}{40}
\expandafter\ifx\csname natexlab\endcsname\relax\def\natexlab#1{#1}\fi

\bibitem[{Bhowmik and de~Melo(2020)}]{bhowmik2020explainable}
Rajarshi Bhowmik and Gerard de~Melo. 2020.
\newblock Explainable link prediction for emerging entities in knowledge
  graphs.
\newblock In \emph{International Semantic Web Conference}, pages 39--55.

\bibitem[{Bordes et~al.(2013)Bordes, Usunier, Garcia-Duran, Weston, and
  Yakhnenko}]{bordes2013translating}
Antoine Bordes, Nicolas Usunier, Alberto Garcia-Duran, Jason Weston, and Oksana
  Yakhnenko. 2013.
\newblock Translating embeddings for modeling multi-relational data.
\newblock In \emph{Neural Information Processing Systems}, pages 2787--2795.

\bibitem[{Boschee et~al.(2015)Boschee, Lautenschlager, O'Brien, Shellman,
  Starz, and Ward}]{DVN/28075_2015}
Elizabeth Boschee, Jennifer Lautenschlager, Sean O'Brien, Steve Shellman, James
  Starz, and Michael Ward. 2015.
\newblock {ICEWS Coded Event Data}.

\bibitem[{Chami et~al.(2020)Chami, Wolf, Juan, Sala, Ravi, and
  R{\'{e}}}]{DBLP:conf/acl/ChamiWJSRR20}
Ines Chami, Adva Wolf, Da{-}Cheng Juan, Frederic Sala, Sujith Ravi, and
  Christopher R{\'{e}}. 2020.
\newblock Low-dimensional hyperbolic knowledge graph embeddings.
\newblock In \emph{Proceedings of the 58th Annual Meeting of the Association
  for Computational Linguistics}, pages 6901--6914.

\bibitem[{Das et~al.(2018)Das, Dhuliawala, Zaheer, Vilnis, Durugkar,
  Krishnamurthy, Smola, and McCallum}]{DBLP:conf/iclr/DasDZVDKSM18}
Rajarshi Das, Shehzaad Dhuliawala, Manzil Zaheer, Luke Vilnis, Ishan Durugkar,
  Akshay Krishnamurthy, Alex Smola, and Andrew McCallum. 2018.
\newblock Go for a walk and arrive at the answer: Reasoning over paths in
  knowledge bases using reinforcement learning.
\newblock In \emph{International Conference on Learning Representations}.

\bibitem[{Dasgupta et~al.(2018)Dasgupta, Ray, and Talukdar}]{dasgupta2018hyte}
Shib~Sankar Dasgupta, Swayambhu~Nath Ray, and Partha Talukdar. 2018.
\newblock {HyTE}: Hyperplane-based temporally aware knowledge graph embedding.
\newblock In \emph{Proceedings of the 2018 Conference on Empirical Methods in
  Natural Language Processing}, pages 2001--2011.

\bibitem[{Dettmers et~al.(2018)Dettmers, Minervini, Stenetorp, and
  Riedel}]{dettmers2018convolutional}
Tim Dettmers, Pasquale Minervini, Pontus Stenetorp, and Sebastian Riedel. 2018.
\newblock Convolutional 2{D} knowledge graph embeddings.
\newblock In \emph{Thirty-Second {AAAI} Conference on Artificial Intelligence},
  pages 1811--1818.

\bibitem[{Ding et~al.(2021)Ding, Han, Ma, and Tresp}]{ding2021temporal}
Zifeng Ding, Zhen Han, Yunpu Ma, and Volker Tresp. 2021.
\newblock Temporal knowledge graph forecasting with neural ode.
\newblock \emph{arXiv preprint arXiv:2101.05151}.

\bibitem[{Garc{\'{\i}}a{-}Dur{\'{a}}n et~al.(2018)Garc{\'{\i}}a{-}Dur{\'{a}}n,
  Dumancic, and Niepert}]{DBLP:conf/emnlp/Garcia-DuranDN18}
Alberto Garc{\'{\i}}a{-}Dur{\'{a}}n, Sebastijan Dumancic, and Mathias Niepert.
  2018.
\newblock Learning sequence encoders for temporal knowledge graph completion.
\newblock In \emph{Proceedings of the 2018 Conference on Empirical Methods in
  Natural Language Processing}, pages 4816--4821.

\bibitem[{Goel et~al.(2020)Goel, Kazemi, Brubaker, and
  Poupart}]{DBLP:conf/aaai/GoelKBP20}
Rishab Goel, Seyed~Mehran Kazemi, Marcus Brubaker, and Pascal Poupart. 2020.
\newblock Diachronic embedding for temporal knowledge graph completion.
\newblock In \emph{Thirty-Fourth {AAAI} Conference on Artificial Intelligence},
  pages 3988--3995.

\bibitem[{Guo et~al.(2019)Guo, Sun, and Hu}]{DBLP:conf/icml/GuoSH19}
Lingbing Guo, Zequn Sun, and Wei Hu. 2019.
\newblock Learning to exploit long-term relational dependencies in knowledge
  graphs.
\newblock In \emph{Proceedings of the 36th International Conference on Machine
  Learning}, pages 2505--2514.

\bibitem[{Guo et~al.(2020)Guo, Zhuang, Qin, Zhu, Xie, Xiong, and
  He}]{guo2020survey}
Qingyu Guo, Fuzhen Zhuang, Chuan Qin, Hengshu Zhu, Xing Xie, Hui Xiong, and
  Qing He. 2020.
\newblock \href {https://doi.org/10.1109/TKDE.2020.3028705} {A survey on
  knowledge graph-based recommender systems}.
\newblock \emph{IEEE Transactions on Knowledge and Data Engineering}.

\bibitem[{Hamilton et~al.(2017)Hamilton, Ying, and
  Leskovec}]{DBLP:conf/nips/HamiltonYL17}
William~L. Hamilton, Zhitao Ying, and Jure Leskovec. 2017.
\newblock Inductive representation learning on large graphs.
\newblock In \emph{Neural Information Processing Systems}, pages 1025--1035.

\bibitem[{Han et~al.(2020{\natexlab{a}})Han, Chen, Ma, and
  Tresp}]{DBLP:conf/emnlp/HanCMT20}
Zhen Han, Peng Chen, Yunpu Ma, and Volker Tresp. 2020{\natexlab{a}}.
\newblock {DyERNIE}: Dynamic evolution of riemannian manifold embeddings for
  temporal knowledge graph completion.
\newblock In \emph{Proceedings of the 2020 Conference on Empirical Methods in
  Natural Language Processing}, pages 7301--7316.

\bibitem[{Han et~al.(2021)Han, Chen, Ma, and Tresp}]{han2021explainable}
Zhen Han, Peng Chen, Yunpu Ma, and Volker Tresp. 2021.
\newblock Explainable subgraph reasoning for forecasting on temporal knowledge
  graphs.
\newblock In \emph{International Conference on Learning Representations}.

\bibitem[{Han et~al.(2020{\natexlab{b}})Han, Ma, Wang, G{\"u}nnemann, and
  Tresp}]{han2020graph}
Zhen Han, Yunpu Ma, Yuyi Wang, Stephan G{\"u}nnemann, and Volker Tresp.
  2020{\natexlab{b}}.
\newblock Graph hawkes neural network for forecasting on temporal knowledge
  graphs.
\newblock In \emph{Conference on Automated Knowledge Base Construction}.

\bibitem[{Jiang et~al.(2016{\natexlab{a}})Jiang, Liu, Ge, Sha, Chang, Li, and
  Sui}]{DBLP:conf/coling/JiangLGSCLS16}
Tingsong Jiang, Tianyu Liu, Tao Ge, Lei Sha, Baobao Chang, Sujian Li, and
  Zhifang Sui. 2016{\natexlab{a}}.
\newblock Towards time-aware knowledge graph completion.
\newblock In \emph{{COLING} 2016, 26th International Conference on
  Computational Linguistics}, pages 1715--1724.

\bibitem[{Jiang et~al.(2016{\natexlab{b}})Jiang, Liu, Ge, Sha, Chang, Li, and
  Sui}]{jiang2016towards}
Tingsong Jiang, Tianyu Liu, Tao Ge, Lei Sha, Baobao Chang, Sujian Li, and
  Zhifang Sui. 2016{\natexlab{b}}.
\newblock Towards time-aware knowledge graph completion.
\newblock In \emph{{COLING} 2016, 26th International Conference on
  Computational Linguistics}, pages 1715--1724.

\bibitem[{Jin et~al.(2020)Jin, Qu, Jin, and Ren}]{jin2020Renet}
Woojeong Jin, Meng Qu, Xisen Jin, and Xiang Ren. 2020.
\newblock Recurrent event network: Autoregressive structure inference over
  temporal knowledge graphs.
\newblock In \emph{Proceedings of the 2020 Conference on Empirical Methods in
  Natural Language Processing}, pages 6669--6683.

\bibitem[{Jung et~al.(2020)Jung, Jung, and Kang}]{jung2020t}
Jaehun Jung, Jinhong Jung, and U~Kang. 2020.
\newblock {T}-{GAP}: Learning to walk across time for temporal knowledge graph
  completion.
\newblock \emph{arXiv preprint arXiv:2012.10595}.

\bibitem[{Lacroix et~al.(2020)Lacroix, Obozinski, and
  Usunier}]{DBLP:conf/iclr/LacroixOU20}
Timoth{\'{e}}e Lacroix, Guillaume Obozinski, and Nicolas Usunier. 2020.
\newblock Tensor decompositions for temporal knowledge base completion.
\newblock In \emph{International Conference on Learning Representations}.

\bibitem[{Leblay and Chekol(2018{\natexlab{a}})}]{LeblayDeriving2018}
Julien Leblay and Melisachew~Wudage Chekol. 2018{\natexlab{a}}.
\newblock Deriving validity time in knowledge graph.
\newblock In \emph{Companion Proceedings of the The Web Conference}, pages
  1771--1776.

\bibitem[{Leblay and Chekol(2018{\natexlab{b}})}]{leblay2018deriving}
Julien Leblay and Melisachew~Wudage Chekol. 2018{\natexlab{b}}.
\newblock Deriving validity time in knowledge graph.
\newblock In \emph{Companion Proceedings of the The Web Conference 2018}, pages
  1771--1776.

\bibitem[{Lin et~al.(2018)Lin, Socher, and Xiong}]{DBLP:conf/emnlp/LinSX18}
Xi~Victoria Lin, Richard Socher, and Caiming Xiong. 2018.
\newblock Multi-hop knowledge graph reasoning with reward shaping.
\newblock In \emph{Proceedings of the 2018 Conference on Empirical Methods in
  Natural Language Processing}, pages 3243--3253.

\bibitem[{Lin et~al.(2015)Lin, Liu, Luan, Sun, Rao, and
  Liu}]{lin-etal-2015-modeling}
Yankai Lin, Zhiyuan Liu, Huanbo Luan, Maosong Sun, Siwei Rao, and Song Liu.
  2015.
\newblock Modeling relation paths for representation learning of knowledge
  bases.
\newblock In \emph{Proceedings of the 2015 Conference on Empirical Methods in
  Natural Language Processing}, pages 705--714.

\bibitem[{Mahdisoltani et~al.(2015)Mahdisoltani, Biega, and
  Suchanek}]{MahdisoltaniBS15}
Farzaneh Mahdisoltani, Joanna Biega, and Fabian~M. Suchanek. 2015.
\newblock {YAGO3}: A knowledge base from multilingual wikipedias.
\newblock In \emph{Seventh Biennial Conference on Innovative Data Systems
  Research}.

\bibitem[{Moon et~al.(2019)Moon, Shah, Kumar, and Subba}]{moon2019opendialkg}
Seungwhan Moon, Pararth Shah, Anuj Kumar, and Rajen Subba. 2019.
\newblock {OpenDialKG}: Explainable conversational reasoning with
  attention-based walks over knowledge graphs.
\newblock In \emph{Proceedings of the 57th Annual Meeting of the Association
  for Computational Linguistics}, pages 845--854.

\bibitem[{Nathani et~al.(2019)Nathani, Chauhan, Sharma, and
  Kaul}]{nathani-etal-2019-learning}
Deepak Nathani, Jatin Chauhan, Charu Sharma, and Manohar Kaul. 2019.
\newblock Learning attention-based embeddings for relation prediction in
  knowledge graphs.
\newblock In \emph{Proceedings of the 57th Annual Meeting of the Association
  for Computational Linguistics}, pages 4710--4723.

\bibitem[{Nickel et~al.(2011)Nickel, Tresp, and Kriegel}]{nickel2011three}
Maximilian Nickel, Volker Tresp, and Hans-Peter Kriegel. 2011.
\newblock A three-way model for collective learning on multi-relational data.
\newblock In \emph{Proceedings of the 28th International Conference on Machine
  Learning}, pages 809--816.

\bibitem[{Schlichtkrull et~al.(2018)Schlichtkrull, Kipf, Bloem, Van Den~Berg,
  Titov, and Welling}]{schlichtkrull2018modeling}
Michael Schlichtkrull, Thomas~N Kipf, Peter Bloem, Rianne Van Den~Berg, Ivan
  Titov, and Max Welling. 2018.
\newblock Modeling relational data with graph convolutional networks.
\newblock In \emph{European Semantic Web Conference}, pages 593--607.

\bibitem[{Sun et~al.(2019)Sun, Deng, Nie, and Tang}]{DBLP:conf/iclr/SunDNT19}
Zhiqing Sun, Zhi{-}Hong Deng, Jian{-}Yun Nie, and Jian Tang. 2019.
\newblock Rotate: Knowledge graph embedding by relational rotation in complex
  space.
\newblock In \emph{International Conference on Learning Representations}.

\bibitem[{Trivedi et~al.(2017)Trivedi, Dai, Wang, and Song}]{trivedi2017know}
Rakshit Trivedi, Hanjun Dai, Yichen Wang, and Le~Song. 2017.
\newblock {Know-Evolve}: Deep temporal reasoning for dynamic knowledge graphs.
\newblock In \emph{Proceedings of the 34th International Conference on Machine
  Learning}, pages 3462--3471.

\bibitem[{Trouillon et~al.(2016)Trouillon, Welbl, Riedel, Gaussier, and
  Bouchard}]{trouillon2016complex}
Th{\'e}o Trouillon, Johannes Welbl, Sebastian Riedel, {\'E}ric Gaussier, and
  Guillaume Bouchard. 2016.
\newblock Complex embeddings for simple link prediction.
\newblock In \emph{Proceedings of the 33nd International Conference on Machine
  Learning}, pages 2071--2080.

\bibitem[{Vashishth et~al.(2020)Vashishth, Sanyal, Nitin, Agrawal, and
  Talukdar}]{vashishth2020interacte}
Shikhar Vashishth, Soumya Sanyal, Vikram Nitin, Nilesh Agrawal, and Partha
  Talukdar. 2020.
\newblock Interact{E}: Improving convolution-based knowledge graph embeddings
  by increasing feature interactions.
\newblock In \emph{Thirty-Fourth {AAAI} Conference on Artificial Intelligence},
  pages 3009--3016.

\bibitem[{Williams(1992)}]{williams1992simple}
Ronald~J Williams. 1992.
\newblock Simple statistical gradient-following algorithms for connectionist
  reinforcement learning.
\newblock \emph{Machine Learning}, 8:229--256.

\bibitem[{Wu et~al.(2020)Wu, Cao, Cheung, and
  Hamilton}]{DBLP:conf/emnlp/WuCCH20}
Jiapeng Wu, Meng Cao, Jackie Chi~Kit Cheung, and William~L. Hamilton. 2020.
\newblock {TeMP}: Temporal message passing for temporal knowledge graph
  completion.
\newblock In \emph{Proceedings of the 2020 Conference on Empirical Methods in
  Natural Language Processing}, pages 5730--5746.

\bibitem[{Xiong et~al.(2017)Xiong, Hoang, and Wang}]{DBLP:conf/emnlp/XiongHW17}
Wenhan Xiong, Thien Hoang, and William~Yang Wang. 2017.
\newblock {DeepPath}: {A} reinforcement learning method for knowledge graph
  reasoning.
\newblock In \emph{Proceedings of the 2017 Conference on Empirical Methods in
  Natural Language Processing}, pages 564--573.

\bibitem[{Zhang et~al.(2018)Zhang, Dai, Kozareva, Smola, and
  Song}]{zhang2018variational}
Yuyu Zhang, Hanjun Dai, Zornitsa Kozareva, Alexander Smola, and Le~Song. 2018.
\newblock Variational reasoning for question answering with knowledge graph.
\newblock In \emph{Thirty-Second {AAAI} Conference on Artificial Intelligence},
  pages 6069--6076.

\bibitem[{Zhang et~al.(2020)Zhang, Zhuang, Zhu, Shi, Xiong, and
  He}]{zhang2020relational}
Zhao Zhang, Fuzhen Zhuang, Hengshu Zhu, Zhiping Shi, Hui Xiong, and Qing He.
  2020.
\newblock Relational graph neural network with hierarchical attention for
  knowledge graph completion.
\newblock In \emph{Thirty-Fourth {AAAI} Conference on Artificial Intelligence},
  pages 9612--9619.

\bibitem[{Zhu et~al.(2021)Zhu, Chen, Fan, Cheng, and
  Zhang}]{zhu-etal-2021-cygnet}
Cunchao Zhu, Muhao Chen, Changjun Fan, Guangquan Cheng, and Yan Zhang. 2021.
\newblock Learning from history: Modeling temporal knowledge graphs with
  sequential copy-generation networks.
\newblock In \emph{Thirty-Fifth {AAAI} Conference on Artificial Intelligence},
  pages 4732--4740.

\end{thebibliography}
\bibliographystyle{acl_natbib}
\clearpage

\appendix
\section{Appendix}

\subsection{Definitions for Evaluation Metrics} \label{Definitions for Evaluation Metrics}
We use two popular metrics,  Mean Reciprocal Rank (MRR) and Hits@$k$ (we let $k$ $\in \{1, 3, 10\}$), to evaluate the models' performance. For each quadruple $q = (e_s, r, e_o, t)$ in the test fact set $\mathcal{F}_{test}$, we evaluate two queries: $q_o = (e_s, r, ?, t)$ and $q_s = (?, r, e_o, t)$. For each query, our model ranks the entities searched by beam search according to the transition probability. If the ground truth entity does not appear in the final searched entity set, we set the rank as the number of entities in the dataset. xERTE ranks the entities in the final inference graph, and others rank the total entities in the dataset. 

MRR is defined as:
\begin{equation}
\begin{split}
    MRR = \frac{1}{2 * |\mathcal{F}_{test}|} \sum_{q\in \mathcal{F}_{test}} (\frac{1}{rank(e_o|q_o)} \\ + \frac{1}{rank(e_s|q_s)}).
\end{split}
\end{equation}

Hits@$k$ is the percentage of times that the ground truth entity appears in the top $k$ of the ranked candidates, defined as:
\begin{equation}
\begin{split}
    \text{Hits}@k = \frac{1}{2 * |\mathcal{F}_{test}|} \sum_{q \in \mathcal{F}_{test}} ( \mathbb{\uppercase\expandafter{\romannumeral1}}\{rank(e_o|q_o\} \leq k \\ + \mathbb{\uppercase\expandafter{\romannumeral1}}\{rank(e_s|q_s\} \leq k).
\end{split}
\end{equation}

There are two filtering settings, static filtering and time-aware filtering. RE-NET~\cite{jin2020Renet} and CyGNet~\cite{zhu-etal-2021-cygnet} directly use static filtering setting to remove the entities from the candidate list according to the triples in the dataset. However, this filtering setting is not appropriate for temporal KGs. The facts of temporal KGs always change over time. For example, we evaluate the test quadruple ($A$, visit, $B$, 2018/04/03). There are two other facts ($A$, visit, $C$, 2018/04/03) and ($A$, visit, $D$, 2018/03/28). Static filtering setting will remove both $C$ and $D$ from the candidates even though the triple ($A$, visit, $D$) is not invalid in 2018/04/03. A more appropriate setting is only to remove $C$ from the candidates. Therefore, we use the time-aware filtering setting that eliminates the entities according to the quadruple.

\subsection{Dataset Statistics} \label{Dataset Statistics}
Dataset statistics are provided in Table \ref{table:dataset statistics}. $N_{train}$, $N_{valid}$ and $N_{test}$ are the numbers of quadruples in training set, valid set, and test set, respectively. $N_{ent}$ and $N_{rel}$ are the numbers of total entities and total relations. ICEWS14 and ICEWS18 are event-based knowledge graphs, and we use the same version as ~\cite{han2021explainable}\footnote{https://github.com/TemporalKGTeam/xERTE}. WIKI and YAGO datasets contain temporal facts with time span $(e_s, r, e_o, [t_s, t_e])$, and each fact is converted to $\{(e_s, r, e_o, t_s), (e_s, r, e_o, t_s + 1_t),...,(e_s, r, e_o, t_e)\}$ where $1_t$ is a unit time. Here, $1_t == 1 year$.  We use the same version of WIKI and YAGO as ~\cite{jin2020Renet}\footnote{https://github.com/INK-USC/RE-Net}.

Since we split each dataset by timestamps, some entities in the test set do not exist in the training set. Table \ref{table: new entities} describes the number of unseen entities and the number of quadruples containing these entities in the test set. $N_{ent}^{unseen}$ is the number of new entities in the test set. $N_{oq}^{unseen}$ is the number of quadruples that object entities are unseen. $N_{sq}^{unseen}$ is the number of quadruples that subject entities are unseen. $N_{soq}^{unseen}$ is the number of quadruples that both subject entities and object entities are unseen.

\begin{table}[]
\centering
\caption{Number of unseen entities in the test set.}
\label{table: new entities}
\resizebox{\linewidth}{!}{
\begin{tabular}{lllll}
\hline
Dataset  & $N_{ent}^{unseen}$  &  $N_{oq}^{unseen}$  &  $N_{sq}^{unseen}$  & $N_{soq}^{unseen}$    \\ \hline
ICEWS14  & 496  &  438  &  497  & 73  \\
ICEWS18  & 1140  &  975  &  1050  & 77    \\
WIKI  & 2968  &  11086  &  22967  & 6974   \\
YAGO  & 540  &  1102  &  873  & 366   \\
\hline
\end{tabular}
}
\end{table}

\begin{table*}[]
\centering
\caption{Statistics on datasets.}
\label{table:dataset statistics}
\resizebox{0.8\linewidth}{!}{
\begin{tabular}{lllllll}
\hline
Dataset              & $N_{train}$  &  $N_{valid}$  &  $N_{test}$  & $N_{ent}$  &  $N_{rel}$ &  Time granularity  \\ \hline
ICEWS14  & 63685  &  13823  &  13222  & 7128  &  230 &  24 hours  \\
ICEWS18  & 373018  &  45995  &  49545  & 23033  &  256 &  24 hours  \\
WIKI  & 539286  &  67538  &  63110  & 12554  &  24 &  1 year  \\
YAGO  & 161540  &  19523  &  20026  & 10623  &  10 &  1 year  \\
\hline
\end{tabular}
}
\end{table*}

\subsection{Detailed Implementation} \label{Detailed Implementation}

\paragraph{Hyperparameters search.} We use a grid search to choose the hyperparameters. The search space of $\mu$ is $[0.1, 0.3, 0.5, 0.7, 0.9]$. The search space of the outgoing edges number $N$ is $[30, 50, 80, 100, 200]$. The search space of the path length is $[2, 3, 4]$. We also try using different activation functions for the $\boldsymbol{\Phi}$, such as $relu, sigmoid, tanh$.

\paragraph{Details of TITer.}  For the policy network, we set the entity embedding dimension to 80, and the function $\boldsymbol{\Phi}$ output dimension to 20. The node representation is the concatenation of the entity embedding and $\boldsymbol{\Phi}$ output, and its dimension is set to 100. we also set the relation embedding size to 100. The hidden state dimension of LSTM and the middle layer dimensions of the two two-layer MLPs are 100. We choose the latest $N$ outgoing edges for the agent as the current state's candidate actions. $N$ is 50 for ICEWS14 and ICEWS18, 60 for WIKI, and 30 for YAGO. The reasoning path length is set to 3. All the parameters are initialized with Xavier initialization.

\paragraph{Details of Training.} Same as MINERVA, we use an additive control variate baseline to reduce the variance and add an entropy regularization term to the cost function scaled by a constant to encourage diversity in the paths sampled by the policy. The scaling constant is initialized to 0.01, and it will decay exponentially with the number of training epochs. The attenuation coefficient is 0.9. The discount factor $\gamma$ of REINFORCE is 0.95. We use Adam optimizer to optimize the parameters, and the learning rate is 0.001. The weight decay of the optimizer is 0.000001. We clip gradients greater than 10 to avoid the gradient explosion. The batch size is set to 512.

\paragraph{Details of Testing.} We use beam search to obtain a list of predicted entities with the corresponding scores. The beam size is set to 100. Multiple paths obtained through beam search may lead to the same target entity, and we keep the highest path score among them as the final entity score. For the IM module, we set the time decay factor $\mu$ to 0.1.

\paragraph{Details of other Methods.} We use the released code to implement DE-SimplE\footnote{https://github.com/BorealisAI/de-simple}, TNTComplEx\footnote{https://github.com/facebookresearch/tkbc}, CyGNet\footnote{https://github.com/CunchaoZ/CyGNet}, RE-NET\footnote{https://github.com/INK-USC/RE-Net}, and xERTE\footnote{https://github.com/TemporalKGTeam/xERTE}. We use the default parameters in the code. Partial results in Table \ref{table:experiment_results_tdf} are from~\cite{han2021explainable, ding2021temporal}. The authors of CyGNet only made object predictions when evaluating their model. We find that the subject prediction is more difficult than the object prediction for these four datasets. We use the code of CyGNet to train two models to predict object and subject, respectively. TANGO~\cite{ding2021temporal} does not release the code, so we use the results reported in their paper.

\subsection{Model Robustness}

We run TITer on all datasets five times by using five different random seeds with fixed hyperparameters. Table \ref{table:robustness} reports the mean and standard deviation of TITer on these datasets. It shows that TITer demonstrates a small standard deviation, which indicates its robustness.

\begin{table*}[]
\centering
\caption{Mean and standard deviation of TITer across five runs on four datasets.}
\label{table:robustness}
\begin{tabular}{lcccc}
\hline
\multicolumn{1}{c}{Datasets} & MRR            & H@1            & H@3            & H@10           \\ \hline
ICEWS14                      & 41.75$\pm$0.21 & 32.75$\pm$0.12 & 46.46$\pm$0.09 & 58.44$\pm$0.05 \\
ICEWS18                      & 29.98$\pm$0.15 & 22.05$\pm$0.07 & 33.46$\pm$0.06 & 44.83$\pm$0.03 \\
WIKI                         & 75.50$\pm$0.22 & 72.96$\pm$0.18 & 77.46$\pm$0.09 & 79.02$\pm$0.04 \\
YAGO                         & 87.47$\pm$0.08 & 84.89$\pm$0.07 & 89.96$\pm$0.03 & 90.27$\pm$0.04 \\ \hline
\end{tabular}
\end{table*}

\subsection{Estimating a Dirichlet Distribution} \label{estimating a Dirichlet distribution}
The Dirichlet density is:
\begin{equation}
    p(\textbf{p}) \sim Dir(\alpha_1,...,\alpha_k) = \frac{\Gamma (\sum_k \alpha_k)}{\prod_k\Gamma(\alpha_k)}\prod_k p_k^{\alpha_k-1},
\end{equation}
where $p_k > 0$, $\sum_k p_k = 1$ and $\alpha_k > 0$. To estimate a Dirichlet distribution of order $k$ with parameters $\bm{\alpha} = \{\alpha_1, \alpha_2, ..., \alpha_k\}$, we observe a set of samples $D = \{x_1,...,x_N\}$ where $x$ is a multinomial sample with length $n$ with probability $\textbf{p}$. $n_k$ represents the count of corresponding category.
\begin{equation}
\begin{split}
    p(\textbf{x}|\bm{\alpha}) &= \int_\textbf{p} p(\textbf{x}|\textbf{p})p(\textbf{p}|\bm{\alpha}) \\
     &= \frac{\Gamma(\sum_k \alpha_k)}{\Gamma(n + \sum_k \alpha_k)} \prod_k \frac{\Gamma(n_k + \alpha_k)}{\Gamma(\alpha_k)},
\end{split}
\end{equation}

\begin{equation}
\begin{split}
    p(D|\bm{\alpha}) &= \prod_i p(x_i|\bm{\alpha}) \\ 
    &= \prod_i (\frac{\Gamma(\sum_k \alpha_k)}{\Gamma(n_i + \sum_k \alpha_k)} \prod_k \frac{\Gamma(n_{ik} + \alpha_k)}{\Gamma(\alpha_k)}).
\end{split}
\end{equation}

The gradient of the log-likelihood is:
\begin{equation}
\begin{split}
   \frac{ d\, log(p(D|\bm{\alpha}) } {d\, \alpha_k} = \sum_i \Psi(\sum_k\alpha_k) - \Psi(n_i+\sum_k\alpha_k) \\ + \Psi(n_{ik}+\alpha_k) - \Psi(\alpha_k),
\end{split}
\end{equation}
where $\Psi$ is the digamma function. The maximum can be computed via the fixed-point iteration:
\begin{equation}
\begin{split}
    \alpha_k^{new} = \alpha_k \frac{\sum_i \Psi(n_{ik}+\alpha_k) - \Psi(\alpha_k)}{\sum_i \Psi(n_i+\sum_k\alpha_k)-\Psi(\sum_k\alpha_k)}.
\end{split}
\end{equation}
We can also maximize the leave-one-out likelihood. The digamma function $\Psi$ can also be inverted efficiently by using a Newton-Raphson method.

\end{document}